\documentclass[sn-mathphys]{sn-jnl}

\jyear{2026}%

\theoremstyle{thmstyleone}%

\theoremstyle{thmstyletwo}%

\theoremstyle{thmstylethree}%

\begin{document}
	
\title[Glance, Scrutinize, and Think]{Glance, Scrutinize, and Think: Advancing Video Anomaly Detection from Training-Free to Agentic Reasoning}
	
\author[1,2]{\fnm{Shibo} \sur{Gao}}
	
\author*[2,3]{\fnm{Peipei} \sur{Yang}}\email{ppyang@nlpr.ia.ac.cn}
	
\author[2,3]{\fnm{Xu-Yao} \sur{Zhang}}
	
\author[1]{\fnm{Linlin} \sur{Huang}}
	
\affil[1]{\orgdiv{School of Electronic and Information Engineering}, \orgname{Beijing Jiaotong University}, \orgaddress{\city{Beijing} \postcode{100044},  \country{China}}}
	
\affil[2]{\orgname{Institute of Automation, Chinese Academy of Sciences}, \orgaddress{\city{Beijing} \postcode{100190},  \country{China}}}
	
\affil[3]{\orgdiv{State Key Laboratory of Multimodal Artificial Intelligence Systems}, \orgname{Institute of Automation, Chinese Academy of Sciences}, \orgaddress{\city{Beijing} \postcode{100190},  \country{China}}}
	
\abstract{Video Anomaly Detection (VAD) aims to identify anomalous events in videos and accurately localize their temporal intervals. Existing approaches exhibit a ``when--what'' dissociation: traditional DNN-based methods localize \emph{when} anomalies occur but lack semantic understanding, whereas emerging LLM-based methods explain \emph{what} happens but neglect precise temporal grounding. This dissociation stems from the absence of a unified reasoning paradigm. Inspired by how humans inspect surveillance videos --- \emph{glancing} globally to form temporal hypotheses, \emph{scrutinizing} suspicious segments, and \emph{thinking} iteratively to correct errors --- we study this global-to-local paradigm from two complementary perspectives. We first propose Glance then Scrutinize (GtS), a training-free framework that leverages static and dynamic textual guidance for coarse-to-fine anomaly grounding and understanding, balancing accuracy and speed. To break the ceiling imposed by frozen external modules, we further propose a tool-augmented agentic VAD method, where a multimodal large language model learns to invoke a video cropping tool, inspect densely resampled frames, and self-correct mislocalized hypotheses, via cold-start supervised fine-tuning followed by reinforcement learning with a joint answer--grounding reward. For training and evaluation, we extend our prior VAGU benchmark into VAGU-T (Video Anomaly Grounding, Understanding, and Thinking), comprising 7,567 real-world videos over 21 anomaly categories with human-validated grounding, semantic explanations, QA pairs, and chain-of-thought tool-calling traces. We further introduce JeAUG, a metric that jointly evaluates semantic interpretability and temporal precision. Extensive experiments show that GtS substantially surpasses existing training-free baselines, while the agentic model delivers both higher accuracy and faster inference, demonstrating the effectiveness and complementarity of the two methods.}
	
	
\maketitle
	
\section{Introduction}\label{sec1}
	
Video Anomaly Detection (VAD) aims to understand and temporally ground anomalous events in video sequences. In recent years, driven by the growing demand for real-time monitoring in industrial automation, intelligent surveillance, smart transportation systems, and even automated livestock monitoring~\cite{PigTracking}, VAD has emerged as a critical research frontier in computer vision and multimedia analytics~\cite{VADSurvey}.

\begin{figure}[t]
	\centering
	\includegraphics[width=\linewidth]{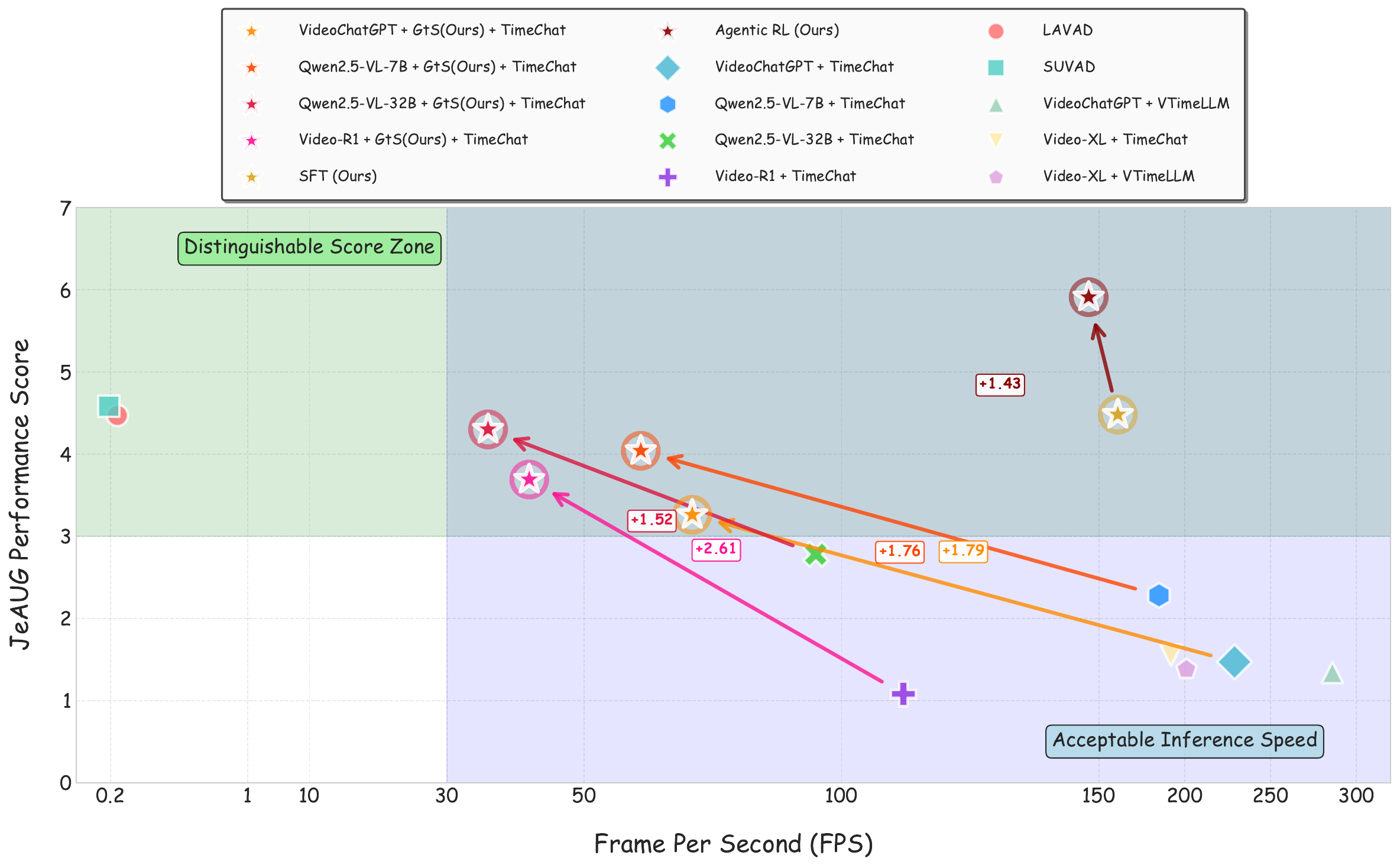}
	\caption{Performance (JeAUG) versus inference speed (FPS) comparison on the VAGU-T suite. The green band marks the distinguishable score zone ($\mathrm{JeAUG} \geq 3$) and the blue band marks the acceptable inference speed ($\mathrm{FPS} \geq 30$). Arrows indicate the improvement of GtS over the corresponding direct VQA/VTG baselines, and of Agentic RL over the SFT model. Our trained agentic model lies in the upper-right region, delivering both high accuracy and real-time speed}
	\label{fig:teaser}
\end{figure}

The current video anomaly detection domain exhibits significant methodological bifurcation: traditional DNN-based methods (spanning semi-supervised, weakly-supervised, and open-set settings)~\cite{learn-28-lu2020few,ECCV_2022_Jigsaw_Puzzles,weakly_Dual_2023,vadclip_vadclip,open-set-UBnormal} and LLM-based methods~\cite{CUVA,HAWK,VERA,Holmes-VAD,VAD-LLaMA} respectively focus on temporal grounding and semantic understanding of anomalies, creating a "when-what" capability dissociation. On one hand, traditional DNN-based methods learn normal/anomalous patterns in video sequences through video-level classification labels to capture anomalous events, yet they merely output temporal grounding results ("when anomalies occur") while lacking semantic understanding. On the other hand, emerging LLM-based methods can generate natural language descriptions leveraging LLMs' open-domain knowledge to answer "what anomalies occur", but universally neglect precise temporal grounding of anomaly onset/offset boundaries. Although some methods attempt joint grounding and understanding through vision-language models (VLMs) via frame/segment-wise video analysis~\cite{LAVAD,SUVAD}, their prohibitively high computational overhead renders them unsuitable for meeting VAD's stringent real-time processing requirements.

We observe that this "when-what" dissociation fundamentally stems from the absence of a unified reasoning paradigm that naturally bridges temporal localization and semantic comprehension. Notably, humans inherently adopt a global-to-local strategy when monitoring surveillance footage: a security operator first performs a rapid scan over the entire video to identify suspicious moments, then zooms into those segments for detailed inspection, and iteratively refines the judgment if the initial assessment proves incorrect. This coarse-to-fine cognitive process seamlessly integrates "when" and "what" — the temporal hypothesis guides where to look, while the semantic analysis validates or revises that hypothesis. However, existing VAD methods fail to replicate this unified reasoning loop: DNN-based methods lack the semantic understanding to form meaningful hypotheses, while LLM-based methods lack the temporal awareness to know where to look.

Inspired by this human cognitive process, we investigate the joint anomaly grounding and understanding problem through a unified global-to-local framework, approaching it from two complementary perspectives:

\textbf{Perspective I: Training-Free Global-to-Local Reasoning.} 
We first ask: can we realize the global-to-local paradigm without any model training? This is motivated by practical considerations — a training-free solution can be deployed immediately with off-the-shelf models, naturally generalizes across scenarios without domain-specific fitting, and serves as a rapid validation of the global-to-local paradigm itself. To this end, we propose \textit{Glance then Scrutinize} (GtS), a training-free framework that leverages static and dynamic textual guidance to first coarsely localize high-probability anomalous regions, then performs detailed anomaly interpretation and temporal boundary refinement. GtS achieves an effective balance between detection accuracy and inference speed without requiring any domain-specific training.

\textbf{Perspective II: Tool-Augmented Agentic Reasoning.} While GtS demonstrates the effectiveness of global-to-local reasoning, its performance is inherently bounded by the capabilities of frozen external modules (e.g., CLIP encoders~\cite{CLIP-radford2021learning} and fixed LLM prompting). When anomalous cues are subtle or videos are extremely long, static textual guidance struggles to achieve precise localization. This motivates our second approach: can we train a model to \textit{internalize} the global-to-local reasoning capability? Inspired by recent progress in tool-augmented video reasoning~\cite{VITAL,LongVT}, we propose a tool-augmented agentic VAD method, where a multimodal large language model learns to autonomously invoke a native video cropping tool to zoom into suspicious temporal segments, perform fine-grained anomaly analysis on densely resampled frames, and self-correct when initial localization proves inaccurate. This is achieved through a two-stage training strategy combining cold-start supervised fine-tuning with reinforcement learning guided by a joint answer-temporal grounding reward.
	
To support both training and evaluation of the above methods, we construct VAGU-T (Video Anomaly Grounding, Understanding, and Thinking), which significantly extends our prior VAGU benchmark~\cite{VAGU} by incorporating tool-augmented chain-of-thought reasoning traces, forming a large-scale data suite that integrates anomaly grounding, anomaly understanding, and agentic training within a unified framework. VAGU-T comprises 7,567 real-world videos spanning 21 major anomaly categories (covering human criminal activities, natural disasters, animal-related injuries, traffic accidents, etc.), with an average length of 2,716 frames. Each instance is annotated with anomaly category, semantic explanation, and precise temporal grounding through rigorous human-in-the-loop validation. In addition, we provide over 20,000 anomaly-related QA pairs to facilitate comprehensive anomaly understanding and objective evaluation. Beyond serving as an evaluation benchmark, VAGU-T further provides tool-augmented chain-of-thought training data — including single-turn and multi-turn reasoning traces with native video cropping tool invocations — that directly supports cold-start SFT and reinforcement learning for our agentic method.

Furthermore, we note that existing VAD evaluation metrics suffer from single-dimensional assessment — either focusing on semantic similarity (ROUGE~\cite{ROUGE}, BLEU~\cite{BLEU}, METEOR~\cite{METEOR}) or temporal precision (AUC, AP) in isolation. To address this, we introduce the JeAUG metric, which jointly quantifies semantic accuracy and grounding precision while incorporating video duration as a weighting factor. This enables more equitable evaluation of VAD capabilities across diverse data scenarios compared to conventional evaluation systems.

\begin{figure}[t]
	\centering
	\includegraphics[width=\linewidth]{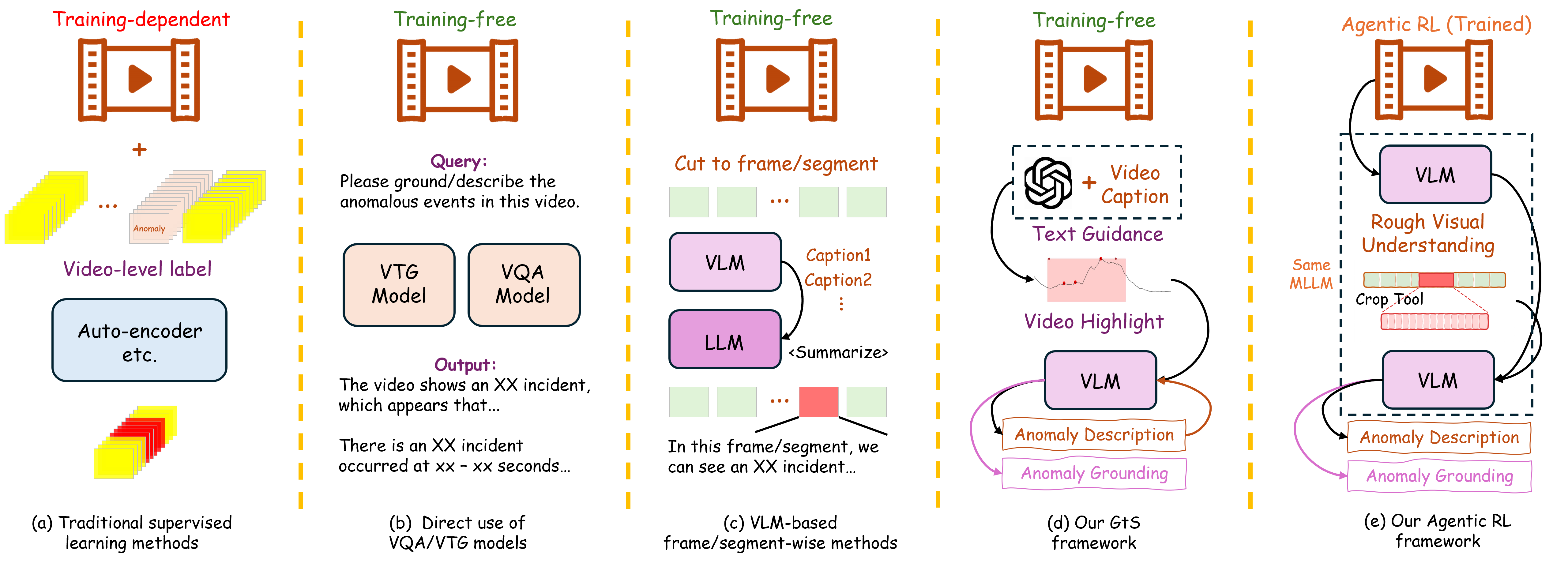}
	\caption{Comparison of VAD paradigms. (a) Traditional supervised methods rely on video-level labels and output temporal localization only (``when''). (b) Direct use of off-the-shelf VQA/VTG models processes the whole video in a single pass. (c) VLM-based frame/segment-wise methods exhaustively caption every segment, incurring prohibitive costs. (d) Our training-free GtS assembles external modules into a glance-then-scrutinize pipeline driven by textual guidance. (e) Our agentic method internalizes the global-to-local reasoning loop into a single MLLM, which natively invokes the crop tool for dense re-inspection and self-corrects mislocalized hypotheses during inference}
	\label{fig:paradigm}
\end{figure}

Fig.~\ref{fig:teaser} illustrates performance and inference speed comparisons between our methods and existing approaches, while Fig.~\ref{fig:paradigm} delineates distinctions between our methods and other VAD paradigms. Extensive experiments on VAGU-T demonstrate that our tool-augmented agentic method significantly outperforms training-free approaches including GtS, which in turn substantially surpasses existing baselines, validating the effectiveness and complementarity of both proposed methods.

Overall, our contributions are summarized as follows:
\begin{itemize}
	\item We propose GtS, a training-free VAD framework that realizes the global-to-local paradigm through static and dynamic textual guidance, achieving an effective balance between detection accuracy and inference speed without any model training.
	\item We propose a tool-augmented agentic VAD method that internalizes the global-to-local reasoning loop into a single multimodal large language model through learned native tool calling, supported by a dedicated training data construction pipeline and a two-stage cold-start SFT and reinforcement learning strategy.
	\item We construct VAGU-T, a large-scale VAD data suite that integrates fully human-validated grounding, understanding, and QA annotations for evaluation with chain-of-thought tool-calling traces for agentic training within a unified framework.
	\item We introduce JeAUG, an evaluation metric that jointly quantifies semantic interpretability and temporal precision, providing more comprehensive and equitable assessment than traditional single-dimensional metrics.
\end{itemize}

A preliminary version of this work, which introduced the VAGU benchmark, the GtS framework, and the JeAUG metric, has been published in our earlier conference paper~\cite{VAGU}.
The present article substantially extends the conference version: we newly propose the tool-augmented agentic VAD method together with its two-stage training strategy (Sec.~\ref{sec:agentic}), enrich the original benchmark into the VAGU-T data suite by constructing chain-of-thought tool-calling traces for agentic training (Sec.~\ref{sec:dataset}), and present extensive new experiments and ablations covering the trained models, reward designs, and efficiency analyses (Sec.~\ref{sec:exp}).

\section{Related Work}
\subsection{Traditional DNN-based VAD Paradigms}
Semi-supervised and weakly-supervised methods remain dominant in video anomaly detection (VAD)~\cite{VADSurvey}. Semi-supervised approaches use self-supervised tasks to learn normal patterns~\cite{learn-28-lu2020few,learn-5-dong2020dual,learn-58-2017,synsyn-astrid2021synthetic,fu-35-wu2010chaotic}, such as auto-encoder reconstruction error and temporal prediction models~\cite{wang2019gods,wu2019deep}, and further improve the efficiency of spatio-temporal modeling through compact convolutional designs~\cite{SemiVAD3D}. Some recent works further improve robustness via multi-task learning~\cite{ECCV_2022_Jigsaw_Puzzles,paper3}, but still struggle with adapting to new scenarios, as minor viewpoint changes can reduce performance.

Weakly-supervised methods use video-level annotations, often with multiple instance learning or semantic priors for anomaly inference~\cite{weakly_Dual_2023,weakly_Generative_2022,vadclip_tsa,vadclip_vadclip}. While these improve detection accuracy, their reliance on manual annotation limits scalability in real-world deployments.

Unsupervised and open-set VAD methods~\cite{open-set-UBnormal,open-set-ding2022catching,open-set-zhu2022towards,open-set-zhu2023anomaly} seek to reduce annotation needs, but often perform poorly across varying scenarios due to architectural limitations. Across all these paradigms, a key challenge remains: limited anomaly understanding, which restricts deployment in complex real-world settings.

\subsection{LLM-Based VAD Paradigms}
Recently, LLM-based VAD methods have advanced significantly. Vision-language models (VLMs) combine LLM reasoning with visual feature extraction, showing strong potential for VAD. Current approaches fall into two main categories:

The first uses external LLMs with frozen VLMs. Videos are segmented, described by VLMs, and processed by LLMs for semantic integration and detection~\cite{CUVA,HAWK,VERA,VAD-LLaMA,VANE-Bench,LAVAD,SUVAD,AnyAnomaly,FollowTheRules}. LAVAD~\cite{LAVAD} provides a training-free pipeline, while SUVAD~\cite{SUVAD} and FtR~\cite{FollowTheRules} introduce rule-mining and hallucination mitigation. Although these methods offer precise grounding and anomaly understanding, their frame/segment-wise processing incurs high computational costs, limiting real-time applications; complementary efforts alleviate the deployment cost of large models through compression, quantization, and distillation~\cite{LBLLM,song2025achieving}, which is orthogonal to the reasoning-level designs studied here.

The second direction enhances VLMs via instruction tuning for interpretable predictions. Holmes-VAD~\cite{Holmes-VAD} uses QA datasets and temporal sampling. VAD-LLaMA~\cite{VAD-LLaMA} improves contextual modeling with LTC modules and a three-stage training strategy. CUVA~\cite{CUVA} adds a MIST selector for feature extraction, and VERA~\cite{VERA} generates instructional questions under weak supervision. While these methods boost detection performance, they require large domain-specific datasets for fine-tuning, leading to high computational cost and limited generalization.

Despite significant progress, existing LLM-based VAD methods share a common limitation: they treat video understanding as a single-pass process, processing the entire video (or uniformly sampled frames) in one forward pass without the ability to adaptively re-examine specific temporal regions. This contrasts sharply with the iterative, evidence-seeking behavior that characterizes effective anomaly analysis in practice.

\subsection{Tool-Augmented Reasoning for Video}
Complementing the development of LLM-based understanding, a recent line of research explores tool-augmented reasoning, where models learn to invoke external tools during inference to enhance their perception and decision-making. In the image domain, methods such as PixelReasoner~\cite{PixelReasoner} and DeepEyes~\cite{DeepEyes} interleave pixel-level operations (e.g., zooming in, drawing auxiliary lines) with reasoning to capture finer details while reducing hallucinations. MERMAID~\cite{MERMAID} further introduces multi-perspective self-reflective agents with generative augmentation for fine-grained visual understanding. For videos, VITAL~\cite{VITAL} demonstrates that equipping multimodal models with a native video cropping tool and training via reinforcement learning significantly improves both video QA and temporal grounding accuracy. Concurrently, VideoThinker~\cite{VideoThinker} proposes reinforcement learning to spark "thinking with videos", enabling models to iteratively refine temporal hypotheses. LongVT~\cite{LongVT} further introduces a three-stage training pipeline (SFT, RL, and RFT) that elicits native tool-calling behaviors for hours-long video reasoning. More recently, EgoR1~\cite{EgoR1} extends tool-augmented reasoning to ultra-long egocentric videos, while ToolAugST~\cite{ToolAugST} equips MLLMs with a comprehensive video toolkit covering temporal grounding, OCR, and frame selection. These approaches share a common philosophy: rather than passively consuming all frames in a single pass, models actively decide what to re-examine and when to seek additional evidence.

However, existing tool-augmented methods are designed for general video question answering, where explicit textual queries guide the model's search. In the VAD setting, no such query is available — the model must autonomously identify what constitutes an anomaly and where to look, making the problem fundamentally more open-ended and challenging. Our work bridges this gap by adapting tool-augmented reasoning specifically for the VAD domain, where the model learns anomaly-aware global-to-local inspection without relying on predefined questions.

\subsection{RL for Multimodal Reasoning}
Reinforcement learning (RL) has recently emerged as a powerful paradigm for enhancing reasoning capabilities in large language models. Inspired by approaches such as DeepSeekR1~\cite{DeepSeekR1}, which demonstrates that RL can incentivize complex reasoning behaviors in text-only models, researchers have extended this paradigm to multimodal domains. In vision-language settings, Group Relative Policy Optimization (GRPO) \cite{GRPO} has been adopted to improve reasoning for image QA \cite{VisionR1,MMEureka} and visual grounding \cite{VLMR1}, and has been further extended with multi-group advantage estimation to accommodate multiple reward signals \cite{MultiGRPO}. A comprehensive survey of this emerging field can be found in \cite{RLMLLMSurvey}.

For video understanding, VideoR1~\cite{VideoR1} applies GRPO-style RL with a temporal-aware T-GRPO algorithm to enhance temporal reasoning in multimodal models. VideoRFT~\cite{VideoRFT} introduces reinforced fine-tuning to incentivize video reasoning without requiring explicit chain-of-thought supervision. TimeR1~\cite{TimeR1} specifically targets temporal video grounding via post-training with IoU-based rewards, while VideoChatR1~\cite{VideoChatR1} enhances spatio-temporal perception through reinforcement fine-tuning. ScalingRLLongVideo~\cite{ScalingRLLongVideo} demonstrates that RL can be effectively scaled to long videos, achieving improved temporal coverage and reasoning completeness. These works collectively demonstrate that RL can push model performance beyond the supervised fine-tuning ceiling by encouraging exploratory reasoning strategies.

Our work leverages this RL paradigm specifically for VAD, designing a joint answer--temporal grounding reward that simultaneously supervises anomaly understanding and localization. While joint rewards have been explored in query-guided settings, the VAD scenario is fundamentally query-free: the reward must incentivize the model to autonomously discover what is anomalous and where it occurs without any textual hint, which imposes a stricter coupling between semantic discovery and temporal search than in general video QA.

\section{Proposed VAGU-T Data Suite}\label{sec:dataset}
In this section, we present the VAGU-T (Video Anomaly Grounding, Understanding, and Thinking) data suite, which significantly extends our prior VAGU benchmark~\cite{VAGU} by incorporating tool-augmented chain-of-thought reasoning traces. VAGU-T serves a dual role: as an evaluation benchmark for assessing joint anomaly grounding and understanding capabilities, and as a training resource providing tool-calling reasoning traces for our agentic method. We first introduce the task definitions, then describe the data collection and annotation pipeline, and finally present dataset statistics with comparative analyses.

\subsection{Task Definition}
\textbf{Video Anomaly Understanding.}
This task comprises two objectives: anomaly classification and anomaly understanding. In the anomaly classification subtask, the model is expected to output the category of the anomalous event present in the video, selected from a predefined anomaly category database. In the anomaly understanding subtask, the model must analyze the given video content to comprehensively describe the subject, process, causes, and consequences potentially involved in the anomalous event.

\textbf{Video Anomaly Grounding.}
This task requires the model to detect precise temporal intervals of anomalous events based solely on video data without external semantic information. Two constraints govern this task. First, apart from the predefined anomaly type list, no video-level or fine-grained semantic labels provided in the dataset may be used, which prevents semantic label leakage. Second, the model is allowed to incorporate anomaly descriptions produced by the upstream anomaly understanding task as grounding evidence, enabling the two tasks to reinforce each other.

\textbf{Tool-Augmented Anomaly Reasoning.}
Beyond evaluation-oriented tasks, VAGU-T additionally defines a training-oriented task: tool-augmented anomaly reasoning. In this task, the model is expected to perform iterative global-to-local reasoning by autonomously invoking a video cropping tool to inspect suspicious temporal segments, analyzing the densely resampled frames for anomalous cues, and self-correcting when initial temporal hypotheses prove incorrect. This task is not used for evaluation but provides structured supervision (in the form of chain-of-thought tool-calling traces) for training our agentic method.

\subsection{Data Collection}
Compared to normal videos, those containing anomalous events are exceptionally scarce. We integrated existing datasets (CUVA~\cite{CUVA}, UCF-Crime~\cite{ucf-crime}, and XD-Violence~\cite{xd-violence}) and collected over 12,000 videos potentially containing anomalies from major platforms such as YouTube, BiliBili, and TikTok. Through rigorous analysis and filtering, we curated 7,567 high-quality anomalous videos spanning 21 distinct anomaly categories across domains including human criminal activities, natural disasters, traffic accidents, and animal-inflicted injuries.

\subsection{Manual Annotation}\label{sec:annotation}
\begin{figure*}[t]
	\centering
	\includegraphics[width=\textwidth]{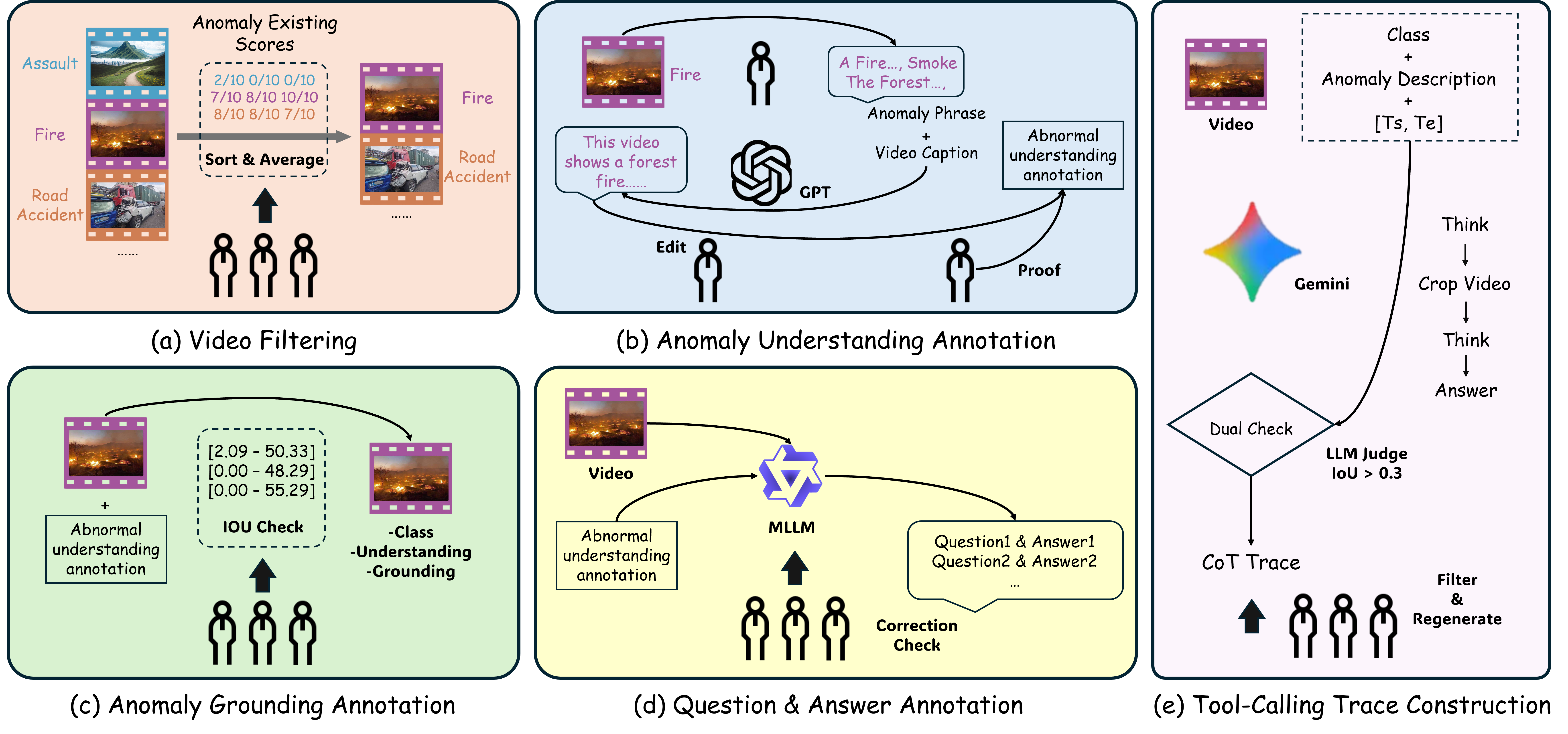}
	\caption{The annotation workflow of the VAGU-T data suite. Stages (a)--(d), inherited from our prior VAGU benchmark~\cite{VAGU}, cover video filtering, anomaly understanding annotation, anomaly grounding annotation, and QA annotation, with each sample processed by at least three annotators collaboratively at every stage. Stage (e) is newly introduced in this work: tool-calling reasoning traces are reverse-engineered from the ground-truth annotations, then retained only after passing the dual check of LLM-as-a-Judge verification and the IoU criterion}
	\label{fig:annotation}
\end{figure*}

\textbf{Basic Data Annotation.}
The construction of our VAGU-T data suite consists of five main steps: video filtering, semi-automatic anomaly annotation, anomaly grounding, QA annotation, and tool-calling trace generation. The first four steps are inherited from our prior VAGU benchmark~\cite{VAGU} and took about 650 hours involving 20 annotators, with at least three annotators working on each video per stage, while the last step is newly introduced in this work and detailed later in this subsection. We collected videos from public datasets and online platforms, removed inappropriate content, and used a three-level quality review to select 7,567 high-quality anomaly videos. For annotation, we adopted a human-AI collaboration: annotators labeled key phrases, generated descriptions with vision-language models, and expanded them with ChatGPT. All annotations underwent multi-person cross-checks for accuracy and consistency. For anomaly grounding, at least three annotators independently marked the time intervals of anomalies, achieving consensus through IoU-based aggregation; ambiguous cases were iteratively re-annotated. We also used advanced multimodal models to create multiple-choice QA tasks for each video, with annotators verifying correctness. Fig.~\ref{fig:annotation} illustrates this process.

\textbf{Tool-Calling Trace Construction.}
To support the training of our agentic method, we further construct tool-augmented reasoning traces based on the human-validated annotations obtained in the previous stages. Specifically, for each annotated video, we have access to three ground-truth elements: the anomaly category, the anomaly description, and the precise anomaly temporal interval $[t_s, t_e]$.

We generate chain-of-thought tool-calling traces that simulate the global-to-local reasoning process using a two-step approach. First, we provide advanced multimodal models (Gemini-2.5-Pro~\cite{Gemini2.5} and GPT-4o~\cite{GPT4o}) with the full video (uniformly sampled at 128 frames) along with the ground-truth annotations, and instruct them to produce a reasoning trace that mimics the human cognitive procedure: the trace first observes the globally sampled frames and forms a temporal hypothesis about where the anomaly likely occurs, then invokes the \texttt{crop\_video}$(t_s, t_e)$ tool to zoom into the hypothesized segment, analyzes the densely resampled frames from the cropped segment to identify and describe the anomaly, and finally outputs the answer containing the anomaly type and description, with the temporal localization given by the window of the final tool call. The model is prompted to "reverse-engineer" a plausible reasoning chain from the known answer, ensuring that each generated trace is both logically coherent and temporally grounded.

Second, to construct multi-turn self-correction traces for longer or more challenging videos, we adopt a length-adaptive strategy inspired by the observation that longer videos are more likely to require iterative refinement. Specifically, we define the probability of selecting a sample for multi-turn trace generation as:

\begin{equation}
	P_{\mathrm{multi}} = \frac{\mathrm{clip}(L_{video},\, L_{min},\, L_{max}) - L_{min}}{L_{max} - L_{min}},
\end{equation}

where $L_{video}$ is the video duration, and $L_{min}$, $L_{max}$ are minimum and maximum duration thresholds. For videos selected under this criterion, we instruct the model to generate traces containing deliberate initial mis-localization followed by self-correction — the model first proposes an incorrect time window, explicitly recognizes that the cropped segment does not contain the target anomaly (e.g., "This segment only shows normal pedestrian activity, not the expected collision"), and then re-invokes the tool with a refined window that correctly covers the annotated ground-truth interval. This design teaches the model the critical skill of error detection and recovery during inference.

Quality control is ensured through a dual-criterion filtering process: a generated trace is retained only if the final answer correctly identifies the anomaly category and provides a semantically consistent description (verified via LLM-as-a-Judge~\cite{GPT4o}), and the temporal interval proposed in the last tool call simultaneously achieves an IoU of at least 0.3 with the ground-truth annotation. Traces failing either criterion are discarded and regenerated. This filtering ensures that only high-quality, well-grounded reasoning traces enter the training pipeline, preventing the model from learning incorrect or hallucinated tool-usage patterns.

\subsection{Dataset Statistics}
The VAGU-T data suite comprises 7,567 high-quality anomaly videos, each annotated with anomaly classification, understanding, and grounding labels. This dataset covers four major domains — human criminal activities, natural disasters, traffic accidents, and animal-related injuries — encompassing 21 fine-grained categories. Following the evaluation protocol of our prior VAGU benchmark~\cite{VAGU}, we select 1,217 samples across the anomaly categories as the held-out evaluation set, while the tool-augmented chain-of-thought reasoning traces are constructed only on the remaining videos, ensuring that no evaluation video is exposed during either SFT or RL training. Table~\ref{tab:benchmark} provides a horizontal comparison with existing datasets through multi-dimensional metrics. Notably, certain categories (e.g., "Fire," "Arson," "Burning") exhibit semantic similarities, and we define explicit differentiation criteria for them during annotation to ensure label consistency.

\begin{table*}
	\caption{A detailed comparison between the proposed VAGU-T data suite and other datasets. A.U. and A.G. mean Anomaly Understanding and Anomaly Grounding. CoT/Tool indicates whether the dataset provides chain-of-thought tool-calling traces for agentic training. The A.G. annotations in the CUVA and HIVAU-70k dataset are accomplished using VLM, which results in significant errors in application.}
	\resizebox{\textwidth}{!}{
		\begin{tabular}{cc|cc|ccccc}
			\toprule
			
			\multirow{2}{*}{Dataset} & \multirow{2}{*}{Domain} & \multicolumn{2}{|c|}{Dataset Statistical Information} & \multicolumn{5}{|c}{Dataset Annotation} \\
			
			\cline{3-9}
			
			& & Video Samples & Anomaly Categories & Audio & A.U. & A.G. & QA & CoT/Tool\\
			
			\midrule
			UCF-Crime & Crime & 1900 & 13 & & & Frame/Human & & \\
			XD-Violence & Violence & 800 & 6 & \checkmark & & Frame/Human & & \\
			ShanghaiTech & Streetscape & 437 & 13 & & & BBox/Human & & \\
			UCSD Ped1 & Streetscape & 70 & 5 & & & BBox/Human & & \\
			UCSD Ped2 & Streetscape & 28 & 5 & & & BBox/Human & & \\
			CUHK Avenue & Streetscape & 37 & 5 & & & BBox/Human & & \\
			Street Scene & Traffic & 81 & 17 & & & BBox/Human & & \\
			CUVA & Multiple & 1000 & 11 & \checkmark & Caption/Human & Period/VLM & & \\
			VANE-Bench & Multiple & 325 & 19 & & & & \checkmark & \\
			HIVAU-70k & Multiple & 5443 & 15 & & Caption/Human & Period/VLM & & \\
			\midrule
			\textbf{VAGU-T (Ours)} & \textbf{Multiple} & \textbf{7567} & \textbf{21} & \textbf{\checkmark} & \textbf{Caption/Human} & \textbf{Period/Human} & \textbf{\checkmark} & \textbf{\checkmark}\\
			
			\bottomrule
		\end{tabular}
	}
	\label{tab:benchmark}
\end{table*}

\section{Method I: GtS — Training-Free Global-to-Local VAD}\label{sec:gts}

\begin{figure}[t]
	\centering
	\includegraphics[width=\linewidth]{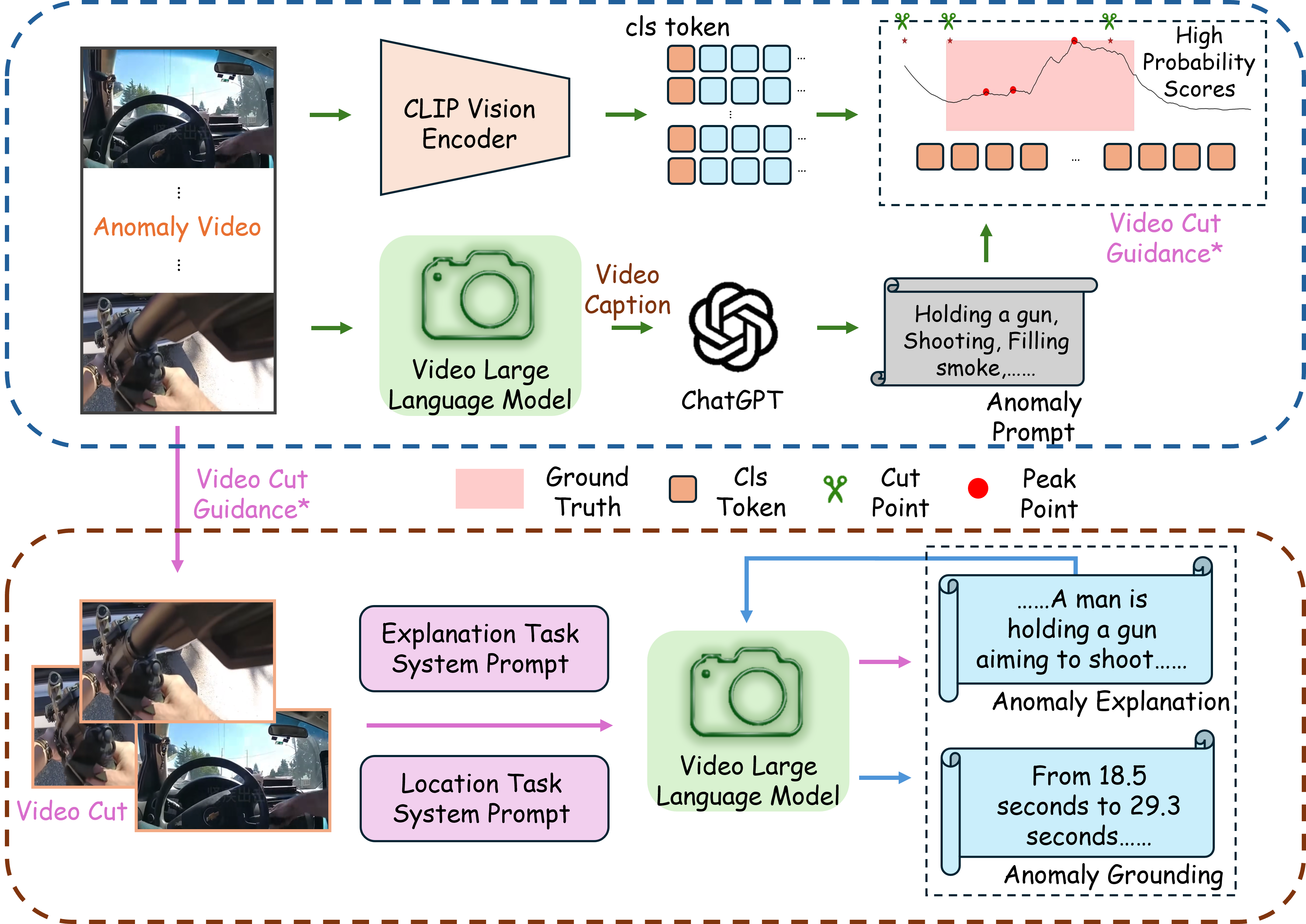}
	\caption{The flowchart of the GtS framework. In the first stage, GtS grounds the time when the main event of the video occurs through dynamic and static text guidance. In the second stage, GtS conducts fine-grained anomaly grounding and understanding based on the video segments obtained in the previous stage}
	\label{fig:model}
\end{figure}

As the first realization of our unified global-to-local paradigm, we propose Glance then Scrutinize (GtS), a training-free VAD framework built upon existing VLMs. GtS implements the coarse-to-fine reasoning process without any model training by employing a dual textual guidance mechanism encompassing both dynamic and static contexts to ground potential anomalous video segments. Since no domain-specific training is required, GtS can be immediately deployed with any compatible VLM, offering fast inference speed that satisfies VAD's real-time requirements while maintaining strong detection accuracy. Fig.~\ref{fig:model} illustrates the workflow of the proposed framework.

\subsection{Glance: Static \& Dynamic Text Guidance and Video Highlight Cut}
Our proposed framework accomplishes video anomaly grounding and understanding through a dual-phase pipeline. During the "Glance" phase, we initially leverage existing VLMs to generate video captions $~Cap_V~$ for input videos $~V_{input}~$. While current VLMs constrained by computational resources often misinterpret genuine anomalies as normal behaviors due to insufficient anomaly-specific token extraction, our experimental findings reveal that the generated descriptions maintain high accuracy in subject recognition. Building upon this insight, we construct multi-source textual inputs incorporating video captions, dataset-provided anomaly lists $~\mathcal{A}~$, and LLM-pregenerated contextual phrase banks $~\mathcal{B}_p~$ to drive GPT in parsing latent dynamic information (actions/events) and static information (subjects/scenes) from videos, thereby generating prompt lists for potential anomaly detection:
\begin{equation}
	\mathcal{PL}_s, \mathcal{PL}_d = {\rm LLM}(~Cap_V~, ~\mathcal{A}~, ~\mathcal{B}_p~).
\end{equation}

Crucially, it is noteworthy that the prompt lists returned by GPT may not inherently contain anomalous content, as this phase fundamentally aims to identify principal video subjects while filtering irrelevant background segments through semantic guidance.

We subsequently employ dual encoders (CLIP $~\Phi_{image}~$ and Video-CLIP $~\Phi_{video}~$) to perform feature encoding on video frames/segments, generating a temporal anomaly probability curve through cross-modal similarity computation:

\begin{equation}
	S_{s/d}(t) = \frac{
		\exp\left( \frac{1}{N} \sum_{n=1}^N \langle \Phi_{x}(pl_x^n),\, \Phi_{x}(V_t) \rangle \right)
	}{
		\sum_{t'} \exp\left( \frac{1}{N} \sum_{n=1}^N \langle \Phi_{x}(pl_x^n),\, \Phi_{x}(V_{t'}) \rangle \right)
	}
\end{equation}
where $x \in \{\mathrm{image}, \mathrm{video}\}$, and $N$ denotes the number of static or dynamic text descriptions in the corresponding branch; $pl_x^n$ represents the static or dynamic text descriptions; $V_t$ refers to the input frame or segment; $\Phi_x$ is the corresponding encoder; and $\langle \cdot, \cdot \rangle$ indicates the cosine similarity.

Considering that events are continuous, we derive the overall similarity curve by combining the two similarity curves and apply the Savitzky-Golay filter to smooth the scores:
\begin{equation}
	S(t) = \frac{1}{Q}\sum_{p=-h}^{+h}\big(\alpha \cdot S_{s}(t+p) + (1-\alpha) \cdot S_{d}(t+p)\big) \cdot q_p,
\end{equation}
where ${q_p}/{Q}$ is the smoothing coefficient, determined through polynomial fitting using the least squares method, $\alpha$ is constant term.

Based on this curve, we implement a three-stage segmentation strategy: firstly detecting local extrema points within the similarity curve, secondly screening top-K candidate peaks according to inter-peak distances and magnitude thresholds:

\begin{equation}
	\mathcal{P} = \{t \in (0,T) \mid S \text{ attains a local maximum at } t\},
\end{equation}
\begin{equation}
	\mathcal{P}^* = \mathrm{TopK}\big(\{p \in \mathcal{P} \mid S(p) \geq \tau\}\big), \quad \text{s.t.}\ \lvert p_i - p_j \rvert > \theta \ \ \forall\, p_i, p_j \in \mathcal{P}^*,
\end{equation}
where $\theta$ and $\tau$ are thresholds.

Finally, we perform dynamic window partitioning around the selected peaks while considering the total video duration, thereby segmenting the original video into high/low anomaly probability segments:
\begin{equation}
	\mathcal{H} = \bigcup_{p \in \mathcal{P}^*} [{\rm max}(0,p-\eta T), {\rm min}(T, p+\eta T)],
\end{equation}
where $\eta \in (0,1)$ controls the window size proportion relative to $T$.

\subsection{Scrutinize: Fine-grained Anomaly Grounding and Understanding}

In the "scrutinize" phase, we establish a closed-loop integration of anomaly understanding and grounding through existing VQA and VTG models. For high-probability anomalous segments, we deploy the VQA model to detect and describe anomalous events based on predefined anomaly catalogs from dataset. 

To more precisely capture anomalous cues, we perform non-uniform sampling on each segmented video clip based on the similarity curve. Specifically, we select frames according to the cumulative distribution of similarity scores, ensuring that the sampling density is proportional to local similarity values. By partitioning the cumulative similarity scores into $N$ equal intervals, we determine the corresponding timestamps as sampling points:

\begin{equation}
	P_i = \min \Big\{ m \in \{a, \ldots, b\} \;\Big\vert\; \sum_{t=a}^{m} S(t) \geq \frac{i}{N} \sum_{t=a}^{b} S(t) \Big\}, \quad i = 1, 2, \ldots, N,
\end{equation}
where $S(t)$ is the similarity score, $\{a, \ldots, b\}$ denotes the segment interval, $P_i$ is the timestamp of the $i$-th sampled frame, and $N$ is the total number of sampled frames.

Furthermore, when performing anomaly detection on each non-initial segment, we provide the model with the understanding results from the preceding segment. This helps to maintain subject consistency in the subsequent integration process.

Compared to holistic video, the segmented analysis significantly enhances VLMs' capability to capture anomaly-relevant tokens, thereby effectively identifying subtle anomalies overlooked in full-length video processing. For low-probability segments, we employ the VQA model to generate captions while extracting latent anomaly-associated clues. 
Subsequently, we utilize a LLM to integrate the aforementioned captions, eliminating repetitive descriptions and those irrelevant to the anomalies, while establishing semantic connections among the segments. 
This process enables our framework to detect causally dependent anomalous behaviors (e.g., theft requiring sequential concealment and escape actions, arson involving combustible material placement and ignition procedures) through multi-segment evidence fusion.

Following comprehensive anomaly characterization, the GtS framework leverages VTG models for temporal grounding. 
By incorporating fine-grained semantic understanding as contextual prompts, our framework achieves superior grounding precision. 
This operational pipeline creates a mutually reinforcing and synergistic relationship between anomaly understanding and anomaly grounding tasks, fundamentally enhancing overall system performance through cognitive-visual alignment.

\section{Method II: Tool-Augmented Agentic VAD}\label{sec:agentic}

However, as a training-free framework, GtS relies entirely on frozen external modules whose capabilities cannot be jointly optimized. When anomalous cues are subtle or videos are extremely long, the static textual guidance may fail to achieve precise localization. To overcome these limitations, we propose a tool-augmented agentic VAD method that trains a unified model to internalize the global-to-local reasoning loop via native tool calling and reinforcement learning.

\subsection{Overview}

Unlike GtS which assembles external modules into a fixed pipeline, our agentic method unifies the entire global-to-local reasoning process within a single multimodal large language model. Given a long video with uniformly sampled frames as global context, the model generates an interleaved chain-of-tool-thought: it first observes the sparse global frames to form a coarse temporal hypothesis about where anomalous events may reside, then invokes a native crop video tool to retrieve densely resampled frames from the hypothesized segment. Based on the retrieved evidence, the model either commits to a final answer containing the anomaly category and description --- with the temporal localization given by the window of its last tool call --- or recognizes that the current segment does not contain the target anomaly and self-corrects by proposing a refined time window for re-inspection. This iterative hypothesis-verification cycle continues until the model is confident in its answer, with a maximum of $T_{max}$ rounds; once the budget is exhausted, the model is required to output its final answer based on the evidence gathered so far.

\begin{figure*}[t]
	\centering
	\includegraphics[width=\textwidth]{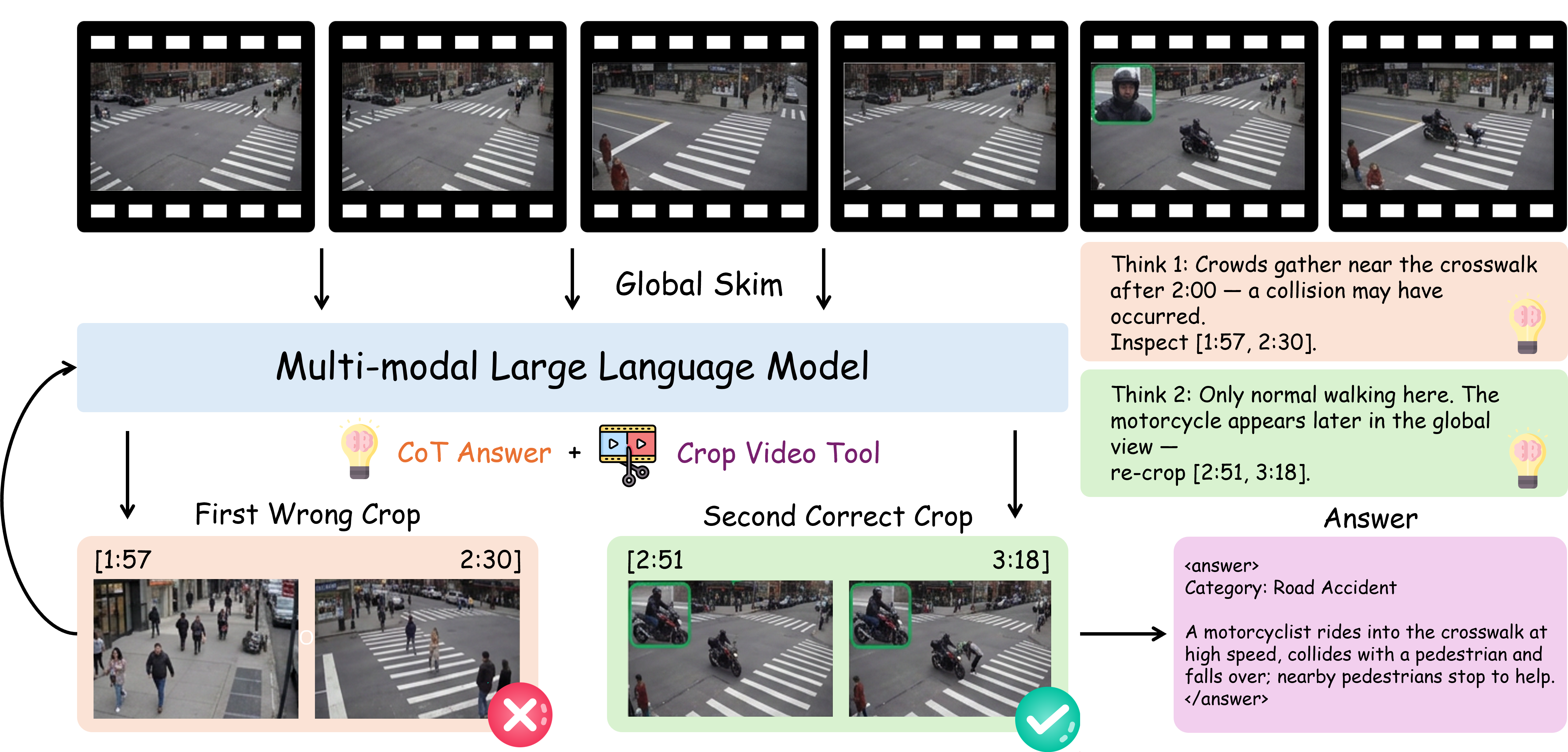}
	\caption{A representative rollout of the proposed tool-augmented agentic VAD method. The MLLM first skims the globally sampled frames and forms a temporal hypothesis (Think 1), then invokes the native crop video tool to inspect the hypothesized segment. Upon recognizing that the first crop contains no anomaly, the model self-corrects (Think 2) and re-invokes the tool with a refined window that captures the motorcycle collision. The final answer contains only the anomaly category and description, while the temporal grounding is given by the window of the last tool call ($[2\!:\!51, 3\!:\!18]$)}
	\label{fig:agentic}
\end{figure*}

A critical distinction from general video QA is that in VAD, no explicit textual query specifies what to search for — the model must autonomously determine what constitutes an anomaly based on visual content alone. This requires anomaly awareness, temporal grounding ability, and evidence integration capability simultaneously. We achieve this through a two-stage training strategy: cold-start supervised fine-tuning establishes basic tool-calling competence and anomaly recognition, followed by reinforcement learning that optimizes exploration and decision-making through outcome-based rewards. Fig.~\ref{fig:agentic} illustrates a representative rollout of the proposed method.

\subsection{Training Data Composition}\label{sec:traindata}

The construction of tool-augmented reasoning traces has been detailed in Sec.~\ref{sec:annotation}, where we described how ground-truth annotations from VAGU-T are leveraged to generate single-turn and multi-turn chain-of-thought tool-calling trajectories using Gemini-2.5-Pro and GPT-4o, along with the length-adaptive multi-turn sampling strategy and dual-criterion quality filtering.

The training corpus is assembled from four complementary sources with differentiated sampling ratios across the SFT and RL stages. The core source is VAGU-T's anomaly-specific tool-call trajectories, which teach the model to construct anomaly-related temporal hypotheses, invoke the cropping tool at appropriate moments, and interpret returned dense frames within the anomaly-detection context; this source constitutes half of the SFT corpus and 70\% of the RL prompts, providing balanced task guidance in the former and intensive policy optimization in the latter. The remaining half of the SFT corpus is evenly drawn from three general sources, each sampled at one third of the VAGU-T trajectory volume so that anomaly-specific and general data maintain a one-to-one ratio. The second source is general video-reasoning data annotated with chain-of-thought reasoning, such as LongVideo-Reason and VideoR1, which preserves broad video understanding and mitigates catastrophic forgetting during domain-specific training. The third source is LLaVA-CoT image-reasoning data, which strengthens foundational visual perception including spatial localization, object recognition, and attribute discrimination—prerequisites for reliable anomaly recognition in video frames. The fourth source is GeminiCoT data provided by LongVT, further extending coverage of long-video chain-of-thought reasoning.

\subsection{Cold-Start SFT}

Our preliminary experiments reveal that directly applying reinforcement learning to the base model leads to training collapse. The model fails to invoke the crop video tool in its generated outputs, and in the rare cases where tool calls appear, it cannot integrate the returned dense frames into coherent anomaly reasoning — often reverting to generic video captioning or repeating earlier observations. This indicates that the base model's latent tool-calling abilities are too weak to support direct RL exploration, making a cold-start SFT stage indispensable.

During SFT, we minimize the standard next-token prediction loss over the assembled training corpus. For a token sequence $x = (x_1, x_2, \ldots, x_L)$ and a model parameterized by $\theta$, the objective is:
\begin{equation}
	\mathcal{L}(\theta) = -\sum_{t=1}^{L} m_t \log p_\theta(x_t \mid x_{<t}),
\end{equation}
where the binary mask $m_t$ restricts the loss to the assistant's output tokens (including think blocks, tool-call invocations, and answer blocks).

After cold-start SFT, the model acquires three fundamental capabilities. It learns to propose plausible temporal windows based on global visual cues — for instance, identifying crowd reactions or scene transitions that suggest anomalous events nearby. It learns to reason over the densely resampled frames returned by the tool, extracting fine-grained anomalous details such as specific actions, object interactions, or environmental changes. And critically, it learns to recognize when its initial temporal hypothesis is incorrect and to propose a revised time window rather than hallucinating an answer from insufficient evidence.

\subsection{Agentic RL}\label{sec:agenticrl}

However, SFT alone remains imitation-driven and suffers from exposure bias: the model fits demonstrated formats but struggles to generalize beyond the training distribution. This motivates the subsequent RL stage.

In this stage, we treat the SFT-trained model as a tool-using agent and employ GRPO~\cite{GRPO} to optimize its reasoning and tool-calling decisions. For each prompt $x$ in the RL prompt set $\mathcal{D}$, we sample $K$ rollout trajectories from the current policy $\pi_{\theta_{old}}$ and compute group-relative advantages:
\begin{equation}
	A^{(k)} = R^{(k)} - \frac{1}{K}\sum_{j=1}^{K} R^{(j)},
\end{equation}
where $R^{(k)}$ is the scalar reward of the $k$-th trajectory. Following Dr.~GRPO~\cite{DrGRPO}, we omit the group standard-deviation normalization, which we find more stable for the long trajectories induced by tool calling.
To obtain a reliable policy gradient in the agentic setting, only the tokens generated by the policy contribute to the objective; the densely resampled frames returned by the crop video tool are masked out, so that the update is computed exclusively over the model's own reasoning, tool-call, and answer tokens. Denoting the sampled trajectory as $o_k$ and the token-level importance ratio as
\begin{equation}
    \rho_{k,t}(\theta) = \frac{\pi_\theta(o_{k,t}\mid x, o_{k,<t})}{\pi_{\theta_{old}}(o_{k,t}\mid x, o_{k,<t})},
\end{equation}
the policy is optimized by maximizing the following PPO-style clipped objective with a KL regularization toward the reference policy $\pi_{ref}$:
\begin{equation}
    \begin{aligned}
        \mathcal{J}(\theta) = \mathbb{E}\Bigg[ \frac{1}{K}\sum_{k=1}^{K}\frac{1}{\lvert o_k\rvert}\sum_{t=1}^{\lvert o_k\rvert} \Big( & \min\big( \rho_{k,t}(\theta)\, A^{(k)},\ \mathrm{clip}(\rho_{k,t}(\theta),\, 1-\epsilon,\, 1+\epsilon)\, A^{(k)} \big) \\
        & - \beta\, \mathbb{D}_{KL}\!\left[\pi_\theta \,\|\, \pi_{ref}\right] \Big) \Bigg],
    \end{aligned}
\end{equation}
where $\epsilon$ is the clipping range that bounds the per-step update, $\beta$ weights the KL penalty, and the KL divergence is estimated with the unbiased estimator
\begin{equation}
    \mathbb{D}_{KL}\!\left[\pi_\theta \,\|\, \pi_{ref}\right] = \frac{\pi_{ref}(o_{k,t}\mid x, o_{k,<t})}{\pi_\theta(o_{k,t}\mid x, o_{k,<t})} - \log \frac{\pi_{ref}(o_{k,t}\mid x, o_{k,<t})}{\pi_\theta(o_{k,t}\mid x, o_{k,<t})} - 1.
\end{equation}

\textbf{Reward Design.}
The scalar reward of each trajectory is the sum of three complementary terms that jointly supervise anomaly understanding, output structure, and temporal grounding:
\begin{equation}
    R = R_{acc} + R_{format} + R_{time}.
\end{equation}
The three terms are weighted equally, reflecting our design principle that a high-quality VAD trajectory must simultaneously produce a semantically correct anomaly interpretation, a well-formed reasoning process, and a temporally precise localization, with none of these objectives dominating the others.

\textit{Accuracy reward $R_{acc}$.} We adopt an LLM-as-a-Judge~\cite{GPT4o} protocol that compares the model's final answer against the ground-truth anomaly description and assigns a three-level score: $R_{acc}=1$ for a fully consistent answer, $R_{acc}=0.5$ for partial consistency, and $R_{acc}=0$ for an inconsistent or empty answer. The judge assesses whether the predicted anomaly category, event description, and causal reasoning match the reference. To prevent the model from hacking the judge by generating overly long, information-padded answers, any answer exceeding a length threshold is assigned $R_{acc}=0$ and simultaneously flagged as a format violation.

\textit{Format reward $R_{format}$.} We grant $R_{format}=1$ only when the output strictly conforms to the interleaved chain-of-tool-thought structure: the sequence must begin with a \texttt{think} block, alternate \texttt{think} and \texttt{tool\_call} blocks in the correct order, and terminate with exactly one \texttt{answer} block; any missing, unbalanced, or misordered tag yields $R_{format}=0$. This term stabilizes training by enforcing a consistently parseable output structure throughout RL exploration.

\textit{Temporal grounding reward $R_{time}$.} We quantify temporal precision by the Intersection-over-Union (IoU) between the model's final localization and the ground-truth anomaly interval $[t_s, t_e]$:
\begin{equation}
    R_{time} = \mathrm{IoU}\big([\hat{t}_s, \hat{t}_e],\, [t_s, t_e]\big) = \frac{\lvert [\hat{t}_s, \hat{t}_e] \cap [t_s, t_e] \rvert}{\lvert [\hat{t}_s, \hat{t}_e] \cup [t_s, t_e] \rvert},
\end{equation}
where $[\hat{t}_s, \hat{t}_e]$ is the temporal window of the \emph{last} crop video invocation in the trajectory, i.e., the segment the model ultimately commits to for fine-grained inspection. If the trajectory contains no tool call, we set $R_{time}=0$, as no explicit localization is produced. Anchoring the reward on the tool window rather than on a self-reported interval ensures that the model is credited only for a localization it has actually inspected, eliminating the possibility of stating a plausible-sounding interval that was never verified. This design also implies that a trajectory must contain at least one tool call to obtain any temporal reward, which further obviates the need for the explicit tool-use bonus discussed below.

\textbf{Why IoU rather than Recall.}
A natural alternative for $R_{time}$ is temporal recall, i.e., the fraction of the ground-truth interval covered by the predicted window. However, recall can be trivially hacked: the model may propose an excessively wide window that fully encloses the anomaly, obtaining perfect recall while providing imprecise boundaries. IoU explicitly penalizes such window inflation through the union term in its denominator, thereby forcing the model to localize both the onset and offset of the anomaly tightly rather than casting an overly broad net.

\textbf{Why no explicit tool-use reward.}
One might additionally reward the model for invoking the crop video tool so as to encourage exploration. We deliberately avoid this. After cold-start SFT, the model has already acquired competent tool-calling behavior; an explicit tool-use bonus would instead incentivize degenerate strategies---calling the tool merely to collect the bonus even when unnecessary---and suppress the model's ability to judge \emph{when} closer inspection is genuinely warranted. We therefore let tool-use behavior emerge implicitly from the outcome-based rewards, so that the model invokes the tool only when doing so improves anomaly understanding and grounding.

\textbf{RL Data Selection.}
We curate the RL prompts from VAGU-T under two considerations. First, we apply difficulty-based filtering: for each prompt we inspect its $K$ sampled rollouts and discard prompts whose rollouts are either all correct or all incorrect, since such prompts yield near-zero group-relative advantages and provide no useful learning signal. Only prompts with mixed outcomes---which produce informative advantage contrasts---are retained. Second, we balance the distribution over video duration to prevent short, easy-to-localize videos from dominating the training batches, ensuring that the model keeps improving on the long-video regime where tool-augmented inspection is most beneficial.

\section{JeAUG: Joint Evaluation of Anomaly Understanding and Grounding}\label{sec:jeaug}

Existing VAD evaluation metrics generally suffer from a single-dimensional assessment limitation. One line of work regards VAD as a video question-answering task and adopts text-similarity measures such as ROUGE~\cite{ROUGE}, BLEU~\cite{BLEU}, and METEOR~\cite{METEOR}, or generation-quality scores derived from GPT-series models~\cite{GPT4o}, thereby focusing solely on the assessment of anomaly semantic understanding. Another line focuses on anomaly spatio-temporal grounding and mainly relies on traditional computer-vision metrics such as AUC and AP. Neither family reflects the fact that a competent VAD system must perform well along both dimensions simultaneously. To this end, we propose the \emph{Joint Evaluation of Anomaly Understanding and Grounding} (JeAUG), a dual-module metric that jointly quantifies semantic accuracy and grounding precision while incorporating video duration as a weighting factor, enabling a more equitable evaluation of VAD capabilities across diverse data scenarios. Overall, JeAUG couples an anomaly-understanding score $\mathrm{Score}_{A.U.}$ with a human-aligned grounding score $F(\mathrm{IoU})$:
\begin{equation}
    \mathrm{JeAUG} = \min\!\left(\gamma\cdot F(\mathrm{IoU}),\, 1\right)\cdot \mathrm{Score}_{A.U.},
\end{equation}
where $\gamma$ is a video-length compensation factor. We detail the two components below.

\subsection{A.U. Score}

In the anomaly-understanding dimension, we aim for models to generate video descriptions that are accurate, clear, concise, and coherent. We therefore guide an external large language model to perform multi-dimensional scoring on the semantic integrity and logical consistency of the generated anomaly descriptions through structured natural-language prompt templates. Specifically, the LLM is prompted to score the response against the ground truth from four aspects---subject, scene, course of events, and impact---on a scale from $1$ to $10$, where $1/10$ indicates that the response is almost entirely unrelated to the ground truth and $10/10$ indicates that it is highly appropriate in every aspect.

Although numerous metrics exist for semantic-similarity assessment (e.g., ROUGE, BLEU, METEOR), descriptions of infrequent anomalous events often vary widely in wording, making these conventional metrics prone to biases arising from sentence length, synonym usage, and word-order variation. To verify the stability of our score, we split each category in the VAGU-T suite into five subsets (rounded down) and compute the Coefficient of Variation (CV) of four metrics; as shown in Table~\ref{tab:cv}, our proposed A.U. score exhibits the greatest stability when processing anomaly-related videos.

\begin{table}[t]
    \centering
    \caption{Comparison of the Coefficient of Variation (CV) for different evaluation metrics. A lower CV indicates greater stability.}
    \label{tab:cv}
    \begin{tabular}{lcccc}
        \hline
         & ROUGE & BLEU & METEOR & JeAUG A.U. \\
        \hline
        CV & 3.12 & 31.72 & 3.33 & \textbf{2.32} \\
        \hline
    \end{tabular}
\end{table}

Furthermore, since JeAUG jointly considers grounding and understanding and the VAGU-T suite supports the integrated application of multiple VLMs, we set the minimum effective score of the anomaly-understanding dimension to $1$ rather than $0$. This avoids assessment imbalances caused by strong performance in one dimension but weak performance in the other (e.g., precise grounding but failed semantic parsing).

\subsection{A.G. Score}

The grounding of anomalous events is highly subjective: in real-world scenarios it is difficult to pinpoint the exact moment an anomaly begins---for a traffic accident, for instance, the starting point may reasonably range from a few seconds to tens of seconds before the collision. To rationalize the grounding metric, we align it with human preferences. We invited more than a dozen evaluators to independently ground anomalies in diverse videos, computed the pairwise Intersection-over-Union (IoU) of their annotations for each video, and averaged the results across all samples as the human-preference consensus, which converges to approximately $0.7$.

\begin{figure*}[t]
    \centering
    \begin{minipage}[t]{0.49\textwidth}
        \centering
        \includegraphics[width=\linewidth]{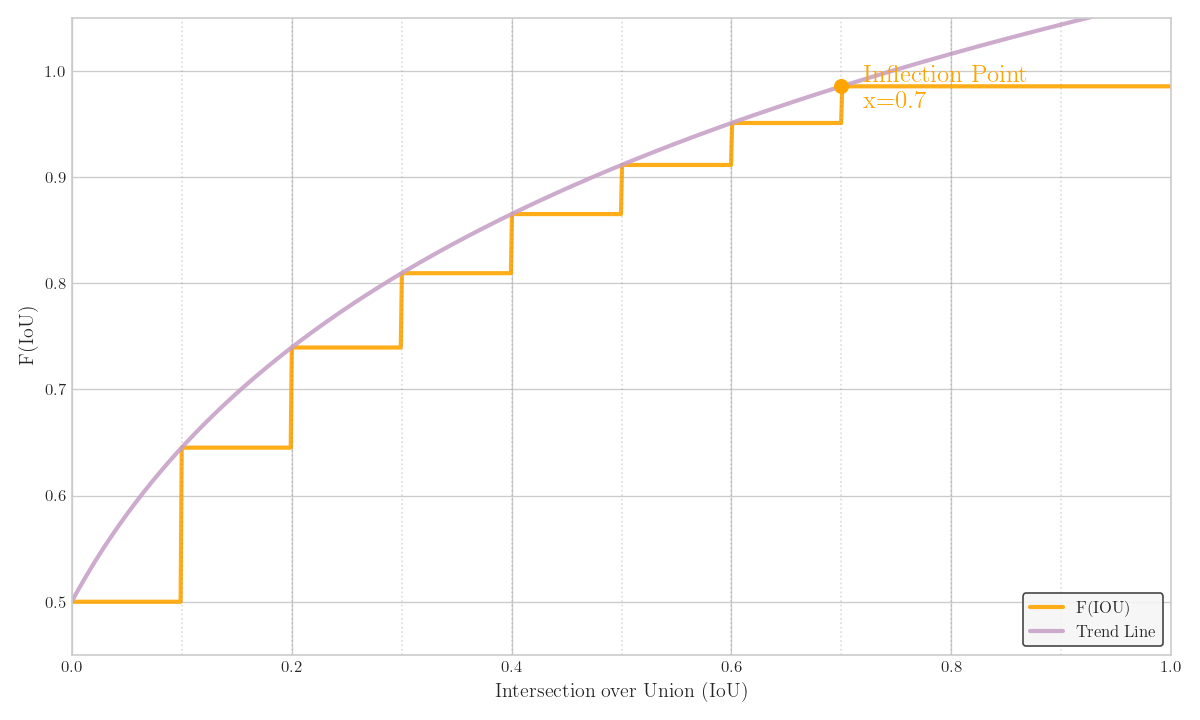}
        \par\smallskip (a)
    \end{minipage}
    \hfill
    \begin{minipage}[t]{0.49\textwidth}
        \centering
        \includegraphics[width=\linewidth]{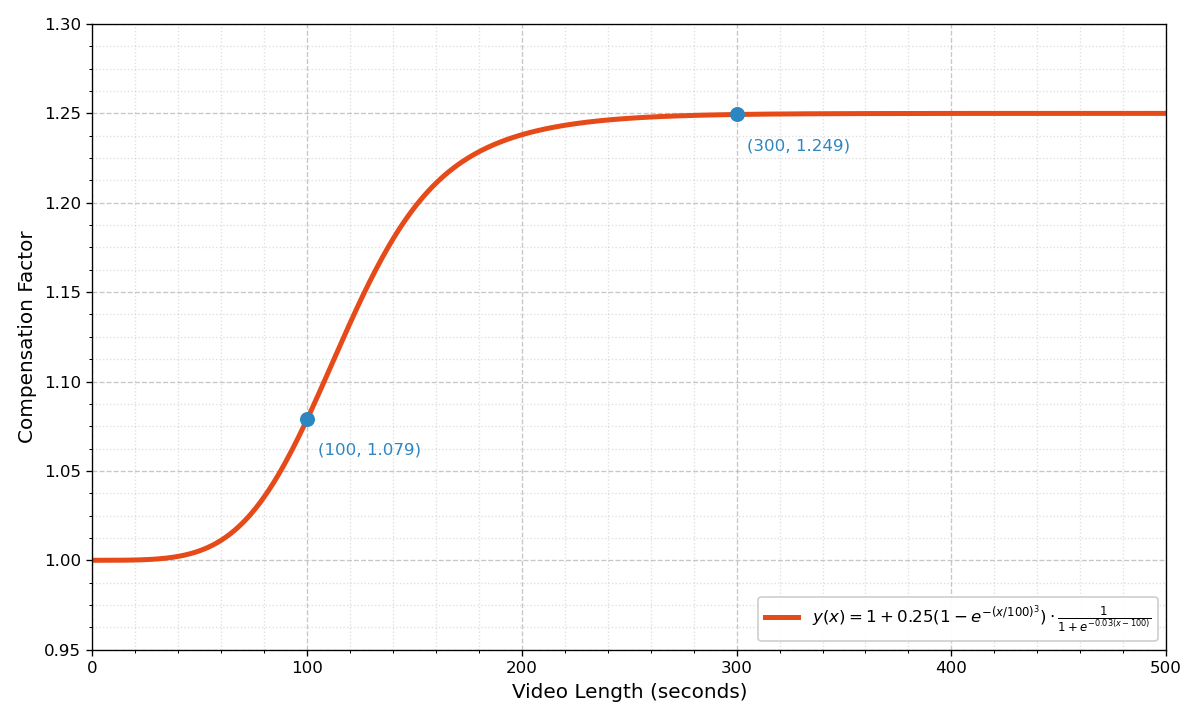}
        \par\smallskip (b)
    \end{minipage}
    \caption{Visualization of the two components of the JeAUG A.G. score. (a) The grounding function $F(\mathrm{IoU})$ starts from $0.5$ to account for failed grounding with successful understanding, increases logarithmically, and saturates once the IoU reaches the human-consensus level of $0.7$. (b) The video-length compensation factor $\gamma$ grows smoothly from $1$ and saturates at $1.25$ for long videos}
    \label{fig:jeaug_curves}
\end{figure*}

Accounting for human grading preferences and the possibility of accurate anomaly comprehension despite failed grounding, we design the following piecewise grounding function:
\begin{equation}
    F(\mathrm{IoU}) = \frac{0.63}{\ln 10}\,\ln\!\left(0.7\cdot\min\!\left(\lfloor 10\cdot\mathrm{IoU}\rfloor,\, 7\right) + 1\right) + 0.5,
\end{equation}
whose design follows four principles: (i) the grounding score reaches its maximum once the IoU attains $0.7$, matching the human consensus; (ii) an initial non-zero score of $0.5$ accounts for cases in which grounding fails but understanding succeeds; (iii) a logarithmic-like increment is adopted so that the score gain gradually slows as the IoU improves, reflecting that coarse-grounding gains (e.g., $0.5\!\rightarrow\!0.6$) matter more than fine ones (e.g., $0.9\!\rightarrow\!1.0$); and (iv) the function is piecewise-defined according to human preference. The visualization of $F(\mathrm{IoU})$ is provided in Fig.~\ref{fig:jeaug_curves}(a).

During our experiments, we further observed that grounding anomalies in longer videos is generally more challenging than in shorter ones: under the premise that general VTG models extract a fixed number of frames, longer videos contain proportionally less anomaly-related information, which increases grounding difficulty. We therefore introduce a video-length compensation factor $\gamma$:
\begin{equation}
    \gamma = G(T) = 1 + 0.25\left(1 - e^{-\left(T/100\right)^3}\right)\left(\frac{1}{1 + e^{-0.03\,(T-100)}}\right),
\end{equation}
where $T$ denotes the video length; its visualization is provided in Fig.~\ref{fig:jeaug_curves}(b).

To motivate this design, we evaluate mainstream VLMs on the anomaly-grounding task under conditions lacking video-level category supervision; as reported in Table~\ref{tab:vtg}, current VLMs perform poorly on anomaly-video grounding. Accordingly, we introduce the length factor $\gamma$ and set $0.5$ as the starting point of $F(\mathrm{IoU})$, which effectively alleviates the evaluation distortion caused by accurate semantic understanding but deviated grounding through a parameterized distribution of task weights.

\begin{table}[t]
    \centering
    \caption{Grounding performance (IoU) of mainstream VLMs on the anomaly-grounding task using the VAGU-T suite.}
    \label{tab:vtg}
    \begin{tabular}{lcccc}
        \hline
        Method & mPLUG-Owl & Video-LLaMA & VideoChatGPT & TimeChat \\
        \hline
        IoU & 12.6\% & 7.3\% & 6.1\% & 14.8\% \\
        \hline
    \end{tabular}
\end{table}

\subsection{Hyperparameters and Distinguishable Thresholds}

The integration of the understanding and grounding branches involves several manually defined hyperparameters, such as the video-length compensation factor $\gamma$ and the weighting coefficients of $F(\mathrm{IoU})$. We emphasize that these hyperparameters were not tuned for any specific dataset; rather, they were set so that the outcomes of JeAUG align as closely as possible with human judgment. To this end, we conducted a human-preference alignment experiment in which $10$ participants independently performed anomaly understanding and grounding on $50$ randomly selected anomalous videos following the same procedure as the MLLM; even the lowest pairwise IoU among participants was approximately $0.7$, corroborating the design choices above.

Finally, to provide interpretable operating points, we define the criteria for acceptable anomaly-understanding ability (A.U.) and anomaly-grounding ability (A.G.) as $\mathrm{JeAUG} \geq 3$ and $\mathrm{FPS} \geq 30$, respectively. The FPS threshold references the frame rates of mainstream video formats and aligns with the standard human viewing speed. Empirically, when the JeAUG score falls below $3$ the model performs poorly in both understanding and grounding anomalous events, whereas above $3$ it generally demonstrates satisfactory performance; we therefore adopt $\mathrm{JeAUG} \geq 3$ as the lower bound of the acceptable range.

\section{Experiments and Analysis}\label{sec:exp}

In this section, we systematically evaluate the two methods proposed in this work---the training-free GtS framework (Sec.~\ref{sec:gts}) and the tool-augmented agentic model (Sec.~\ref{sec:agentic})---on the VAGU-T suite using the JeAUG metric (Sec.~\ref{sec:jeaug}). We first describe the experimental setup, then report the main quantitative comparison against mainstream baselines, followed by ablation studies, in-depth analyses, and qualitative case studies.

\subsection{Experimental Setup}\label{sec:setup}

We evaluate all methods on the VAGU-T suite and report four complementary indicators: the anomaly-understanding score (A.U., $1$--$10$), our joint metric JeAUG, the multiple-choice QA accuracy (QA), and the inference speed (FPS). QA is reported only for the methods whose pipelines natively support the multiple-choice protocol: the frame/segment-wise pipelines output captions and anomaly curves rather than choice selections, while our trained models adopt the open-ended chain-of-tool-thought answer format, and adapting them to the multiple-choice protocol would require dedicated re-training that is orthogonal to the focus of this work.

\textbf{Training-free GtS.} Our GtS framework employs different VLMs as the anomaly-understanding model and the anomaly-grounding model, and uses CLIP-L/14~\cite{CLIP-radford2021learning} to encode video frames. An LLM (Llama-3.1-8B~\cite{Llama3}) is used for caption integration. For the segmentation hyperparameters, we set $\alpha=0.4$; the smoothing coefficient $q_p/Q$ is obtained through least-squares polynomial fitting; $\tau$ is set to the mean of all peak values of the similarity curve; $\theta = \text{(total frame number)}/12$; and $\eta = 1/20$. These hyperparameters were derived by randomly sampling $100$ anomalous videos that were discarded from (i.e., not included in) the evaluation set and computing the proportion of anomalous timestamps. All GtS experiments were conducted on $14$ A6000 GPUs, taking approximately $210$ hours in total.

\textbf{Trained models (SFT \& Agentic RL).} Building upon the two-stage pipeline described in Sec.~\ref{sec:agentic}, we further train a unified tool-augmented model. The cold-start SFT stage and the subsequent GRPO-based agentic RL stage share the training corpus detailed in Sec.~\ref{sec:traindata}. The detailed training configuration, including the base MLLM and the SFT/RL hyperparameters, is summarized in Table~\ref{tab:hparam}.

\begin{table}[t]
    \centering
    \caption{Training configuration of our SFT and Agentic RL models.}
    \label{tab:hparam}
    \begin{tabular}{ll}
        \hline
        Item & Value \\
        \hline
        Base MLLM & Qwen2.5-VL-7B-Instruct \\
        SFT epochs / learning rate & $2$ / $1\times10^{-5}$ (cosine decay) \\
        RL rollouts per prompt $K$ & $8$ \\
        Clipping range $\epsilon$ & $0.2$ \\
        KL coefficient $\beta$ & $1\times10^{-3}$ \\
        Max tool-call rounds $T_{max}$ & $3$ \\
        Global / cropped frames & $128$ / $128$ \\
        Hardware & $24\times$ Alibaba T-Head PPU (96\,GB) \\
        \hline
    \end{tabular}
\end{table}

\subsection{Main Results}

Table~\ref{tab:main} reports the comparison between our methods and mainstream baselines on the VAGU-T suite. The baselines fall into two groups: frame/segment-wise pipelines (LAVAD, SUVAD) that process the video exhaustively, and training-free combinations that directly apply off-the-shelf VQA and VTG models to the VAD task. Our methods occupy three rows: the training-free GtS framework built on different backbones, and the two trained variants (SFT and Agentic RL).

The results show that directly applying VQA/VTG models to VAD performs poorly, as these models struggle to locate the most anomaly-relevant cues within long videos. Frame/segment-wise pipelines achieve strong A.U./JeAUG scores but at prohibitively low inference speed (FPS $<1$), violating VAD's real-time requirement. In contrast, GtS attains a favorable balance: on the same backbone it substantially improves both A.U. and JeAUG over the corresponding direct baseline (e.g., Qwen2.5-VL-7B: JeAUG $2.28\!\rightarrow\!4.04$) while keeping FPS above the acceptable threshold of $30$.

The trained models push the frontier further. The SFT model already surpasses the strongest frame/segment-wise pipeline SUVAD in A.U. ($6.62$ vs.\ $5.73$) while running nearly three orders of magnitude faster, confirming that internalizing the global-to-local reasoning loop eliminates the need for exhaustive segment processing. Agentic RL brings an additional gain of $+0.73$ A.U. and $+1.43$ JeAUG over SFT, demonstrating that outcome-based reward optimization elicits capabilities beyond trace imitation. Overall, the Agentic RL model achieves the best accuracy among all compared methods (A.U. $7.35$, JeAUG $5.91$) at $148$ FPS, exceeding the real-time threshold by a large margin (see also Fig.~\ref{fig:teaser}).

\begin{table}[t]
    \centering
    \caption{Comparison on the VAGU-T suite. A.U. is the anomaly-understanding score, JeAUG is our joint metric, QA is the multiple-choice accuracy (\%), and FPS is the inference speed. TC = TimeChat~\cite{TimeChat}, VT = VTimeLLM~\cite{VTimeLLM}; $^*$ denotes integration with our GtS framework; ``--'' indicates that the multiple-choice QA protocol is not applicable to the method (see Sec.~\ref{sec:setup}).}
    \label{tab:main}
    \begin{tabular}{lcccc}
        \hline
        Method & A.U. & JeAUG & QA (\%) & FPS \\
        \hline
        \multicolumn{5}{l}{\textit{Frame/Segment-wise}} \\
        LAVAD~\cite{LAVAD} & 5.52 & 4.47 & -- & 0.24 \\
        SUVAD~\cite{SUVAD} & 5.73 & 4.58 & -- & 0.19 \\
        \hline
        \multicolumn{5}{l}{\textit{Training-free (direct VQA/VTG)}} \\
        VideoChatGPT~\cite{VideoChatGPT} + TC & 2.32 & 1.47 & 60.3 & 229 \\
        VideoChatGPT + VT & 2.10 & 1.34 & 60.3 & 286 \\
        Video-XL~\cite{Video-XL} + TC & 2.31 & 1.55 & 58.6 & 192 \\
        Video-XL + VT & 2.13 & 1.38 & 58.6 & 201 \\
        Qwen2.5-VL-7B + TC & 3.61 & 2.28 & 68.0 & 185 \\
        Qwen2.5-VL-32B + TC & 4.43 & 2.78 & 71.9 & 95 \\
        Video-R1~\cite{VideoR1} + TC & 1.70 & 1.08 & 80.0 & 112 \\
        \hline
        \multicolumn{5}{l}{\textit{Ours: GtS (training-free)}} \\
        Qwen2.5-VL-7B + TC$^*$ & 5.50 & 4.04 & 73.5 & 61 \\
        Qwen2.5-VL-32B + TC$^*$ & 5.99 & 4.30 & 76.8 & 36 \\
        Video-R1 + TC$^*$ & 5.19 & 3.69 & 88.9 & 42 \\
        VideoChatGPT + TC$^*$ & 4.42 & 3.26 & 65.1 & 71 \\
        \hline
        \multicolumn{5}{l}{\textit{Ours: Trained}} \\
        SFT & 6.62 & 4.48 & -- & 161 \\
        Agentic RL & 7.35 & 5.91 & -- & 148 \\
        \hline
    \end{tabular}
\end{table}

\subsection{Ablation Studies}

\textbf{GtS module ablation.} We investigate the contribution of each component of the GtS framework, with results reported in Table~\ref{tab:gts_ablation}. Removing the dynamic text guidance, the static text guidance, the integral non-uniform sampling, or the cross-segment contextual understanding each degrades the A.U. score, confirming that every module enhances the anomaly-detection capability of GtS.

\begin{table}[t]
    \centering
    \caption{Ablation of the GtS modules on the VAGU-T suite (A.U. score).}
    \label{tab:gts_ablation}
    \begin{tabular}{lc}
        \hline
        Model & A.U. \\
        \hline
        Qwen2.5-VL-7B + GtS & 5.50 \\
        \quad w/o dynamic text guidance & 5.27 \\
        \quad w/o static text guidance & 5.30 \\
        \quad w/o integral non-uniform sampling & 5.38 \\
        \quad w/o cross-segment contextual understanding & 5.41 \\
        \hline
    \end{tabular}
\end{table}

To verify that the improvement of GtS does not merely stem from the increased number of sampled frames induced by video segmentation, we uniformly split each video into seven segments (the maximum number observed for GtS on VAGU-T) and apply a similar processing pipeline. As shown in Table~\ref{tab:sampling}, uniform splitting brings only a marginal gain, whereas GtS yields a much larger improvement; further analysis reveals that uniform segmentation often fragments events and thus aggravates hallucination.

\begin{table}[t]
    \centering
    \caption{Effect of different sampling strategies (A.U. score).}
    \label{tab:sampling}
    \begin{tabular}{lc}
        \hline
        Method & A.U. \\
        \hline
        Qwen2.5-VL-7B + TC & 3.61 \\
        \quad + Uniform Split & 4.02 \\
        \quad + GtS & 5.50 \\
        \hline
    \end{tabular}
\end{table}

\textbf{Training and reward ablation.} For the trained models, we further ablate the training stage (SFT vs.\ SFT+RL) and the reward design ($R_{acc}$, $R_{format}$, $R_{time}$) introduced in Sec.~\ref{sec:agenticrl}. The corresponding results are summarized in Table~\ref{tab:train_ablation}. Adding the RL stage on top of SFT already improves the A.U. score, and further incorporating the temporal grounding reward $R_{time}$ yields the best performance, confirming that jointly supervising anomaly understanding and temporal localization is beneficial.

\begin{table}[t]
    \centering
    \caption{Ablation of the training stages and reward components for the trained model.}
    \label{tab:train_ablation}
    \begin{tabular}{lc}
        \hline
        Configuration & A.U. \\
        \hline
        SFT only & 6.62 \\
        SFT + RL ($R_{acc}+R_{format}$) & 6.93 \\
        SFT + RL ($R_{acc}+R_{format}+R_{time}$) & 7.35 \\
        \hline
    \end{tabular}
\end{table}

\subsection{Analysis}

\begin{figure*}[t]
    \centering
    \includegraphics[width=\textwidth]{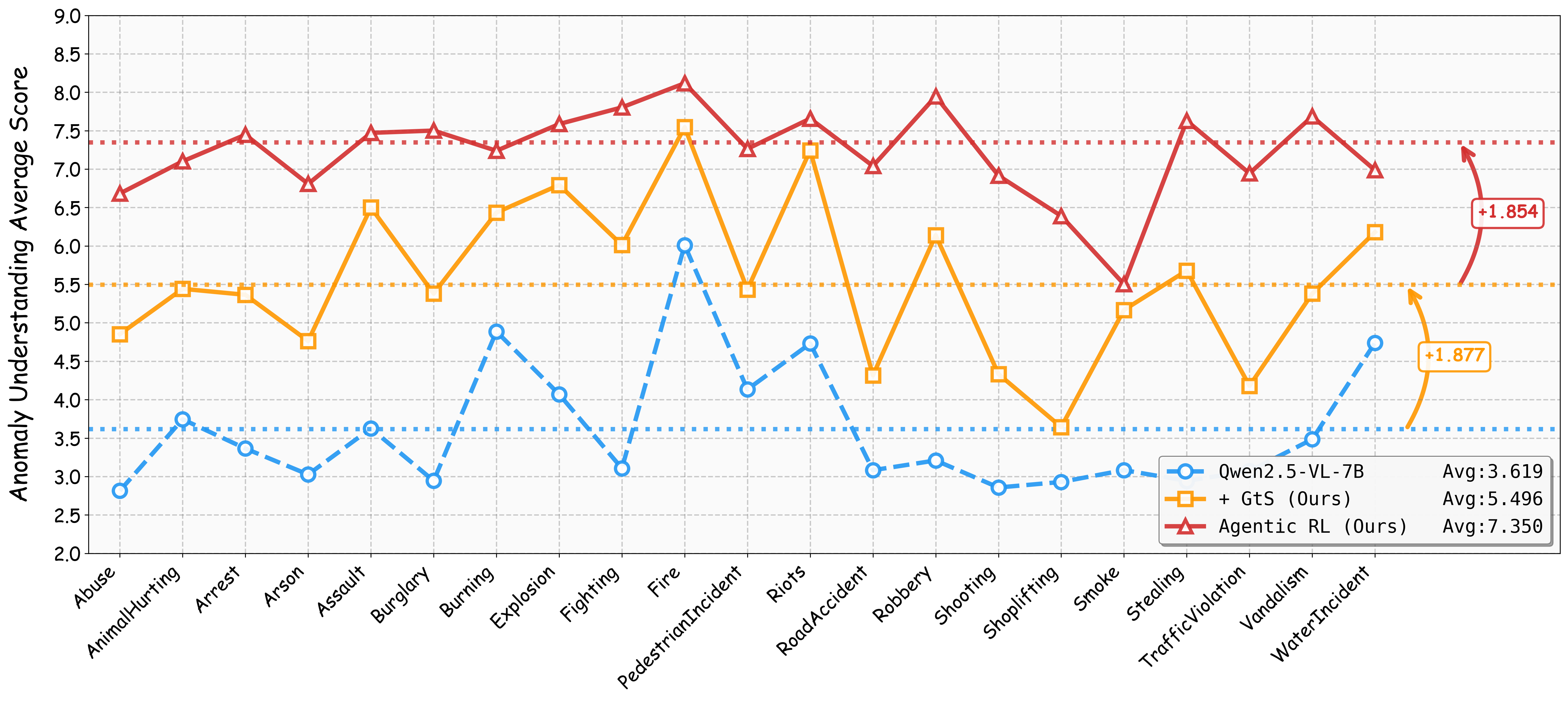}
    \caption{Fine-grained per-category comparison on the video anomaly understanding sub-task. Dotted horizontal lines denote the overall average of each method, and the arrows on the right indicate the average improvement of GtS over the base model and of Agentic RL over GtS}
    \label{fig:category}
\end{figure*}

\textbf{Category-wise analysis.} We conduct a fine-grained per-category analysis on the anomaly-understanding sub-task, with results shown in Fig.~\ref{fig:category}. GtS improves the base model on every one of the $21$ categories, and its gains are most pronounced on anomalies that manifest as salient multi-actor events, such as Robbery ($+2.93$), Fighting ($+2.91$), Assault ($+2.88$) and Riots ($+2.51$), where textual guidance readily localizes the region of interest. Its gains are, however, markedly smaller on categories whose anomalous cues are subtle or extremely short-lived, such as Shoplifting ($+0.71$), TrafficViolation ($+1.11$) and RoadAccident ($+1.23$). Notably, these are exactly the categories on which the Agentic RL model contributes the most over GtS (TrafficViolation $+2.77$, Shoplifting $+2.75$, RoadAccident $+2.72$, Shooting $+2.58$), which empirically confirms our motivation for Method II: static textual guidance is insufficient when anomalous evidence is fine-grained, whereas a learned tool-calling policy can actively zoom in and re-examine such moments. The remaining weak spot is Smoke, whose absolute score stays lowest ($5.50$) as its visual signature is diffuse and it is the rarest category in the suite.

\begin{figure}[t]
    \centering
    \includegraphics[width=\linewidth]{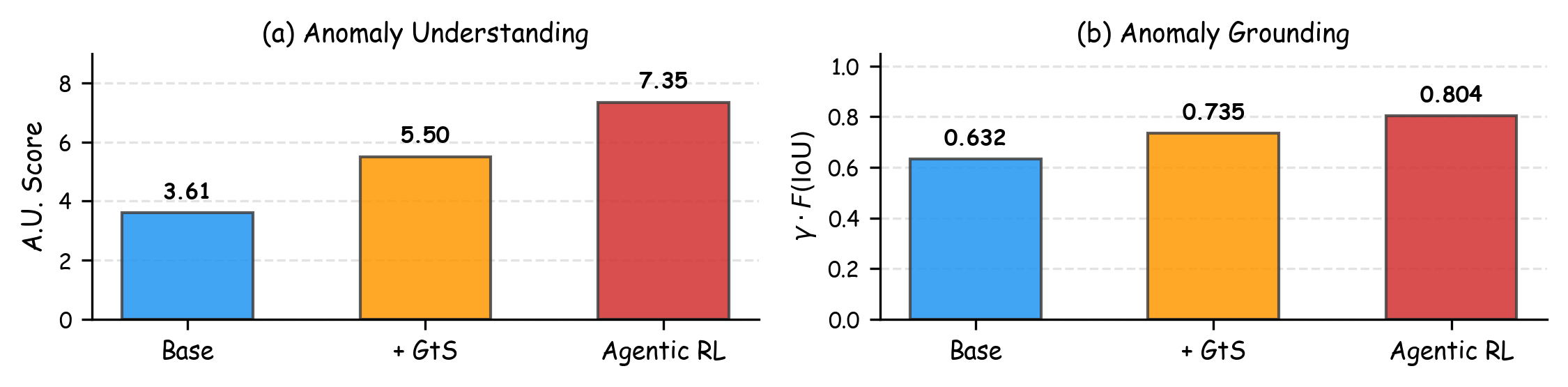}
    \caption{Independent evaluation on the anomaly understanding and anomaly grounding sub-tasks. The grounding factor $\gamma\cdot F(\mathrm{IoU})$ is the grounding component of JeAUG}
    \label{fig:subtask}
\end{figure}

\textbf{Sub-task analysis.} To further dissect performance, we evaluate anomaly understanding and anomaly grounding independently, as illustrated in Fig.~\ref{fig:subtask}; for grounding we adopt the grounding factor $\gamma\cdot F(\mathrm{IoU})$ from JeAUG as the evaluation criterion. Both methods improve on both sub-tasks, but their profiles differ: GtS lifts understanding from $3.61$ to $5.50$ while raising the grounding factor only moderately ($0.632\!\rightarrow\!0.735$), since its temporal boundaries are ultimately produced by a frozen VTG model. The Agentic RL model improves both dimensions further ($7.35$ and $0.804$), with the grounding gain directly attributable to the temporal reward $R_{time}$ that supervises where the model chooses to look. This is consistent with the reward ablation in Table~\ref{tab:train_ablation}, where removing $R_{time}$ costs $0.42$ A.U.

\textbf{Efficiency analysis.} As reported in Table~\ref{tab:main}, frame/segment-wise pipelines are accurate but far below the real-time threshold (FPS $<1$), while direct VQA/VTG models are fast but inaccurate. GtS keeps the FPS above the acceptable threshold of $30$ while markedly improving accuracy. The trained models shift this accuracy-efficiency trade-off even further: by invoking the crop video tool only when closer inspection is warranted rather than processing every segment exhaustively, they sustain around $150$ FPS together with the best accuracy. Interestingly, the SFT model runs slightly faster than the Agentic RL model ($161$ vs.\ $148$ FPS); this gap mainly stems from the SFT model's unstable tool-call format, which occasionally terminates trajectories prematurely, whereas the RL model consistently completes the full inspection loop at a modest cost of roughly $8\%$ inference speed.

\section{Conclusion}

In this paper, we systematically study the joint problem of video anomaly grounding and understanding from a unified global-to-local perspective. To support this study, we introduce VAGU-T, a large-scale data suite that, for the first time, integrates anomaly category, semantic explanation, precise temporal grounding, Video QA, and chain-of-thought tool-calling traces within a single benchmark; compared with existing datasets, VAGU-T is more comprehensive, more challenging, and of higher annotation quality, with all annotations undergoing multiple rounds of manual verification. Building upon this data suite, we investigate two complementary realizations of the global-to-local paradigm. First, we propose Glance then Scrutinize (GtS), a training-free framework that leverages dynamic and static textual guidance to perform coarse-to-fine anomaly grounding and understanding without any model training, achieving a favorable balance between computational cost and detection performance. Second, to overcome the inherent performance ceiling of training-free approaches, we propose a tool-augmented agentic VAD method that internalizes the global-to-local reasoning loop into a single multimodal large language model through a two-stage cold-start SFT and agentic reinforcement learning strategy. Furthermore, we propose the JeAUG metric, which jointly evaluates semantic interpretability and temporal precision and thus overcomes the single-dimensional limitations of traditional metrics such as AUC and AP. Extensive experiments validate the effectiveness and complementarity of both proposed methods. We believe that VAGU-T, together with the two methods and JeAUG, can significantly encourage the exploration of VAD in the era of large language models. In the future, we will further extend the data suite and the agentic framework to more complex and realistic anomaly scenarios.

\section*{Declarations}

\bmhead{Funding}
This work was supported in part by the Key-Area Research and Development Program of Guangdong Province (No.~2024B0101040008).

\bmhead{Competing interests}
The authors have no competing interests to declare that are relevant to the content of this article.

\bmhead{Ethics approval}
This study does not involve experiments on human participants or animals. All videos used in this work were obtained from publicly available academic datasets and public online platforms.

\bmhead{Data availability}
The annotations of the VAGU-T data suite, including the anomaly categories, semantic explanations, temporal grounding labels, QA pairs, and tool-calling reasoning traces, will be made publicly available to the research community upon publication. The source videos are collected from publicly available datasets and online platforms, and are used in accordance with their original licenses.

\bmhead{Author contributions}
Shibo Gao: methodology, software, experiments, and writing of the original draft. Peipei Yang: supervision and writing---review and editing. Xu-Yao Zhang and Linlin Huang: assistance with the study. All authors read and approved the final manuscript.

\bibliography{paper5}
	
\end{document}